\documentclass[10pt]{llncs}

\usepackage{times}
\usepackage{url}
\usepackage{latexsym}
\usepackage[latin9]{inputenc}
\usepackage{listings}
\usepackage{amsmath}
\usepackage{amssymb}
\usepackage{graphicx}
\usepackage{subscript}

\title{Combinatorial Decision Dags: \\ A Natural Computational Model \\ for General Intelligence}

\author {Ben Goertzel}
\institute{SingularityNET Foundation and OpenCog Foundation }

\begin{document}

\maketitle

\begin{abstract}
A novel computational model (CoDD) utilizing combinatory logic to create higher-order decision trees is presented. A theoretical analysis of general intelligence in terms of the formal theory of pattern recognition and pattern formation is outlined, and shown to take especially natural form in the case where patterns are expressed in CoDD language. Relationships between logical entropy and algorithmic information, and Shannon entropy and runtime complexity, are shown to be elucidated by this approach. Extension to the quantum computing case is also briefly discussed.
\end{abstract}

\section{Introduction}

The theoretical foundations of general intelligence outlined in {\it  The Hidden Pattern} \cite{Goertzel2006a} and formalized in earlier works going back to 1991 \cite{Goertzel1991} are fundamentally grounded in the notion of {\it pattern}.   Minds are conceived as patterns emergent in physical cognitive systems, and emergent between these systems and their environments.   Intelligent activity is understood as the process of a system recognizing patterns in itself and its environment,  including patterns in which actions tend to achieve which results in which contexts, and then choosing actions to fit into these recognized patterns.

The formalization of the pattern concept standardly used in this  context is based on algorithmic information  -- in essence a pattern in $x$ is a compressing program for $x$, a program shorter than $x$ that computes $x$.   The definition can be extended to incorporate factors like runtime complexity and lossy compression, but the crux remains algorithmic information.
As program length depends on the assumed  underlying computer,  this formalization approach has an undesirable arbitrariness as its route, though of course as entity sizes go to infinity this arbitrariness becomes irrelevant due to bisimulation arguments.

Here we present a conceptually  cleaner foundation for the pattern-theoretic analysis of general intelligence, in the  form of a new formulation of the pattern  concept in terms of distinctions and decisions.   We present a specific computational model -- Combinatorial Decision Dags or CoDDs -- with a high degree of naturalness in the context of cognitive systems, and use CoDDs to explore the relationship between distinction, pattern, runtime complexity, Shannon entropy and logical entropy.   We also briefly indicate extensions of these ideas to the quantum domain, in which  Boolean distinctions are replaced with amplitude-labeled quantum distinctions and classical patterns are replaced by "quatterns."

The goal is  to provide a clear and  simple mathematical framework that intuitively matches the requirements of general  intelligence, founded on a computational model designed with the requirements of modeling cognitive systems in mind.

\section{Conceptualizing Pattern in Terms of Distinction and Decision}\label{pattern}

Taking our cue from G. Spencer-Brown \cite{SpencerBrown1967}, let us begin our with the
elemental notion of {\it distinction}.

A distinction is a distinction between one collection of entities $A$ and
another collection of entities $B$. Two distinctions are distinguished
from each other if:

\begin{itemize}
\item
  One distinguishes $A_1$ from $B_1$
\item
  The other distinguishes $A_2$ from $B_2$
\item
  It's not the case that $A_1$ and $A_2$ are identical and $B_1$ and $B_2$ are
  identical.
\end{itemize}

Consider a program that takes certain inputs and produces output from  them.  We are then moved to ask: Can we think about the "simplicity" of a process as $\rho$ersely related to the number of distinctions it makes? I.e. is the "complexity" of a
program well conceived in terms of the number of pairs of legal inputs
to which it assigns different outputs?  \footnote{Note that the inputs and outputs of programs may also be programs --
i.e. we can consider ourselves in an "algorithmic chemistry" $T_Y$pe domain \cite{Fontana1991} \cite{Goertzel1994}
comprising a space S of programs that map inputs from S into outputs in
S. This can be formalized in various ways including set theory with an
Anti-Foundation Axiom.)}

This line of thinking meshes naturally with the concept of {\it logical entropy} \cite{ellerman2013introduction}  -- where the logical entropy of a partition of $n$ elements is the percentage of pairs $(x,y)$ of elements so that $x$ and $y$ live in different partition cells.  If we consider a program as a partition of its
inputs, where two inputs go into the same partition cell if they produce
the same output, then the logical entropy of this partition is one
measure of the program's complexity. The simpler programs are then the
ones with the lowest logical entropy.

There is an interesting relationship between program length and
this sort of program logical entropy. For each $N$ there will be some
upper bound to the logical entropy of programs of length $N$, and it's not hard to see that most
programs of length $N$ will have logical entropy fairly near this upper bound.

\subsection{Shannon entropy, program specialization and runtime
complexity}\label{shannon-entropy-program-specialization-and-runtime-complexity}

The relationship between logical entropy and Shannon entropy is also worth exploring.

Consider a case where each possible input for a program is represented
as a (generally long) bit string.

Given the set of distinctions that a program makes (considering the
program as a partition of its inputs), we can also ask: If we were to
effect this set of distinctions via a sequence of distinctions $\rho$olving
individual entries in input bit-strings, how long would the sequence
need to be? E.g. if we break things down into: First distinguish portion
$I_1$ of input space from portion $I_2$ of input space (using a single bit of
the input), then within $I_1$ distinguish subregion $I_{11}$ from subregion $I_{12}$
(using another bit of the input), etc. -- then how many
distinctions need to be in this binary tree of distinctions?

The leaves of this binary decision tree are the partition elements; and
the length of the path from the root to a given leaf, is the number of
binary questions one needs to ask to prove that some input lies in that
particular partition cell.

This decision tree is closely related to the Shannon entropy of the
partition implied by the program. Suppose we have a probability
distribution on the inputs to the program. The Shannon entropy of this
distribution is a lower bound on the average length of the path from the
root of the tree to a leaf of the tree, i.e. a lower bound on the
"average tree depth." (The average is taken over all possible inputs
drawn from the distribution.)

The optimal binary decision tree for the partition, relative to a given
distribution on the bit strings being partitioned, can be considered as
the one for which the average length from root to leaf is minimal. Not
knowing the distribution on inputs, a heuristic is to guess the optimal
tree will be one of those with the fewest nodes.

If we assume a simplified computational model in which one binary
distinction is made per unit time, then the average depth of this binary
decision tree is related to the average runtime complexity of the
program. It tells you how long it would take, on average over all
possible inputs, to run the program on an abstract machine making one
distinction (based on one bit of the input string) per unit time. For
fixed input length $N$, this is, it would seem, a lower bound for the
average runtime complexity of the program on a machine with rapid access
to enough memory to store this huge tree.

Other, more practical instantiations of the same program achieve greater
compactness by $\rho$olving operations other than simply comparing
individual bits of input strings. These other instantiations may run
faster on machines that don't have rapid access to enough memory to
store a huge binary decision tree. They involve "overhead" in the sense
that they use more complex mechanisms to do what could be more simply
done using a series of binary judgments based on input bits; but these
complex mechanisms allow a lot of binary judgments to be carried out
using a smaller amount of memory, which is better in the case of a
computing system that has rapid access only to a relatively small amount
of memory, and much slower access to a larger auxiliary memory.

This binary decision tree can be viewed as a "program specialization" of
the original program to the case of input sequences of length $N$ or less.
Like most program specializations it removes abstraction and creates tremendous
 bloat $\rho$olving a lot of nested conditionals.

Given a program (or a process more generally), then, we can characterize
this program via the set of distinctions it makes between its inputs.
Given a distribution over the inputs, one can calculate the logical
entropy of the set of distinctions (which is a measure of how complex is
the action of the program in terms of its results), and one can also
calculate the average depth of the optimal binary decision tree for
emulating the action of the program, which is a measure of how complex
is the action of the program in terms of its runtime requirements. Of
course one can also quantify the distinction-set implied by the program
in a lot of other ways; these are merely the simplest relevant
quantifications.

Philosophically, if we begin with the partition of input space rather than the program,
we can view the construction of the binary decision tree as a form of
the emergence of time. That is: time arises from the sequencing involved
in constructing the binary decision tree, which intrinsically
incorporates a notion of one decision occurring after another. The
"after" here basically has the semantics "in the context of" -- the
next step is to be interpreted in terms of the previous step, rather
than vice versa. The notion of complex structures unfolding over time
then emerges from the introduction of "memory space" or "size" as a
constraint -- i.e. from the desire to shrink the tree while leaving
the action the same, and increasing the average runtime as little as
possible.

\subsection{Grounding Pattern in
Distinction}\label{grounding-pattern-in-distinction}

A pattern being a "representation as something simpler" -- the core
intuition underlying the classic compression-based conceptualization
of pattern -- can be
formulated in terms of distinction as follows.

Suppose one has

\begin{itemize}
\item
  an "invariant-set", meaning a function $\rho$ that distinguishes certain
  distinctions that are relevant and certain that are not
\item
  a program $F$ that makes certain distinctions among its potential inputs
  (by mapping them into different outputs)
\end{itemize}

Then we may say: \emph{\textbf{$P$ is a pattern in $F$ , relative to $\rho$, if
it makes all the distinctions $F$ does that $\rho$ identifies as relevant,
but makes fewer distinctions than $F$ overall}}...

Extending this, we could say that: $P$ is an approximate pattern in $F$ if:
it makes $K$ fewer distinctions than $F$ overall, and misses fewer than $K$ of
the distinctions $F$ does that $\rho$ identifies as relevant

To incorporate runtime complexity, if we have a weighting on the
distinctions judged relevant by $\rho$, we could require additionally that
\emph{\textbf{the optimal binary decision tree for $P$ has lower average
root-to-leaf length than the optimal binary decision tree for $F$}}
(relative to this weighting). This means that in a certain idealized
sense, $P$ is "fundamentally faster" than $F$.

The degree of runtime optimization provided by $P$ in this sense could be
included in the definition of approximate pattern as a multiplicative
factor.

\section{A Quantum
Definition of
Pattern}\label{one-level-weirder-qudits-and-quatterns-...-a-quantum-definition-of-pattern}

To extend these ideas in the direction of quantum computing, we extend the notion of a {\it dit} (a distinction) to that of a {\it qudit} -- a distinction btw two entities, labeled with a (complex) amplitude.

Assume as above  one has an "invariant-set" defining function $\rho$ that assigns a
``relevance amplitude'' to each qudit in its domain; and assume one is
given a quantum program $F$ whose inputs are vectors of amplitudes, and
that maps each input into different outputs with different
amplitude-weights.

Associate $P$ (for instance) with a "distinction vector" $P^*$ that has
coordinate entries corresponding to pairs of the form {[}input set 1,
input set 2{]} where the entry in the coordinate is the amplitude
assigned to the distinction between the output produced by $P$ on input
set 1 and the output produced by $P$ on input set 2 \ldots{}

Then $P$ is a {\it quattern} in $F$ , relative to $\rho$, if

\begin{itemize}
\item
  \textbar{}\textless{}$\rho$\textgreater{}\textless{}$F^* -
  P^*$\textgreater{}\textbar{} is small
\item
  \textbar{}$P^*$\textbar{} \textless{} \textbar{}$F^*$\textbar{}
\end{itemize}

So one can define a quattern intensity degree via a formula like

(\textbar{}$F^*$\textbar{} - \textbar{}$P^*$\textbar{}) * (
\textbar{}$\rho$\textbar{} -
\textbar{}\textless{}$\rho$\textgreater{}\textless{}$F^*-P^*$\textgreater{}\textbar{}
) / \textbar{}$\rho$\textbar{}

This ends up looking a bit like the good old definition of pattern in
terms of compression, but it's all about counting distinctions (qudits)
now.

A quantum history is a network of interlinked qudits... So a quantum
distinction graph is a network of qudits between quantum distinction
graphs.

Runtime complexity can also be analyzed similarly to in the "classical"
case considered above.

Recall the basic concept of a quantum decision tree \cite{Bera2010}.
In a simple, straightforward formulation, algorithm on inputs of size n
works on 3 registers $I, B, W$ where $I$ has $log(n)$ qubits and is used to
write a query, $B$ has one qubit and is used to store the answer to a
query and $W$  is the workspace register with polynomially many qubits. The
query steps are modeled as particular unitary operators, and the algorithm is allowed to perform intermediate
computations between the queries in the form of unitary operators
independent of the input. 

In this formalism, a $k$-query decision tree $A$ is the unitary
operator $A = U_k O U_{k-1} \dots U_1 O U_0$  and the output of the
algorithm is the value obtained when the first qubit of A\textbar{}0, 0,
0\textgreater{} is measured in any given basis.

The quantum decision tree complexity  $Q\_2$ is the depth of the
lowest-depth quantum decision tree that gives the result $f(x)$ with
probability at least $2/3$ for all $s$. Another quantum decision tree
complexity measure, $Q_E$, is defined as the depth of the lowest-depth
quantum decision tree that gives the result $f(x)$ with probability 1 in
all cases (i.e. computes exactly). Other variations are obviously
possible. These sorts of measures are evidently analogues of the
approach to runtime complexity proposed above. It has been shown that
the Shannon entropy of a random variable computed by the function $f(X)$
is a lower bound for the $Q_E$ quantum decision tree complexity of $f$
cite{shi2000entropy}.

In assessing the degree to which $P$ is a quattern in $F$, one can then look
at the quantum decision tree complexity of $P$ versus that of $F$, similarly
to how in the classical case one looks at the size of the decision tree
associated with $P$ versus that of the decision tree associated with $F$.

Similarly to in the classical case, the runtime complexity measure
depends in a messy but apparently inevitable way on assumptions
regarding the memory of the underlying computing machine. In the quantum
case, if we restrict the workspace $W$ further than just saying it has to
be polynomial in the input size $n$, then we will in generally get larger
decision trees.

Also similarly to the classical case, here real computation usually
involves quantum circuits that do more than just query the input
repeatedly and combine the query results -- thus resulting, much of
the time, in smaller programs that however involve more complex
operators. But the size of the quantum decision tree complexity
summarizes, in a sense, the temporal complexity involved in doing what
the program does, at a basic level, without getting into tricks that may
be used to accelerate it on various computational architectures with
various processing speeds associated with various particularly-sized
memory stores.

\subsection{Weidits and Weitterns}\label{weidits-and-weitterns}

What we have done above with amplitude-valued distinctions, one could do
perfectly well for distinctions labeled with others sorts of weights.
Quaternions e.g. would seem unproblematic, as Banach algebras over
quaternions are understood and relatively well-behaved. The notion of
qudit and quattern in this way can be generalized to weidit and
weittern, defined relative to any sort of weight, not necessarily
complex number weights.

\section{Combinatorial Decision Dags: A Pattern-Based Computational
Model}\label{a-pattern-based-computational-model}

One can also use these ideas to articulate a novel, foundationally pattern-oriented
universal computational model. Of course there are numerous universal
computational models already in practical and theoretical use, but one that is grounded in distinctions and
patterns may be especially useful in a  cognitive modeling and AGI context,  if  one adopts
a view of general intelligence that places pattern at the coore.

It is straightforward to make the decision-tree rendition of programs
recursive -- just take a bit-string encoding of a decision tree and
feed it as input to another decision tree. In this way we create
decision trees that represent higher-order functions. We just need to
add encoder and decoder primitives, mapping back and forth between
decision trees and bit strings, to our basic language.

This leads us to the notion  of Combinatorial Decision Dags (CoDDs).   Defining a $k$'th order decision dag as one that takes $k-1$'st order decision dags as inputs, it is clear that $k$'th order decision trees (or dags) are equivalent to SK combinator expressions, thus have universal expressive capability among Turing computable functions. 

To be a bit more explicit: In this context, programs are viewed roughly as follows:

\begin{itemize}
\item
  Start with (higher order) decision trees, then find cases where there
  is a pattern $P$ in subtree $T_X$relative to subtree $T_Y$.
\item
  Then replace $T_X$with $P$ plus a pointer to $T_Y$ as $P$'s input (this is
  "pattern-based memo-ization")
\item
  Repeat, and eventually one gets a compact, complex program rather than
  a forest of recursively nested decision trees
\end{itemize}

\noindent Here $P$ is of course a function that can be represented as a decision
tree, or else as a decision tree with pattern-based memo-ization as
described above.

The $K$ combinator $K y x = y$ (using curried notation) is a function so that, given any input $y$,
$K y$ is a decision-tree that always outputs $y$. Given the encoding of
(first or higher order) decision trees as bit strings, K combinator for bit string inputs can be applied
to any decision tree as input.

The $S$ combinator has the form $S f x y = (f x) (f y)$.   So if (still currying) we have a tree taking (a bit-string-encoded
version of) another tree as input,

$$
T_{X_1} T_{X_2}
$$

and we then have the same pattern in both $T_1$ and $T_2$,

$$
(P T_{Y_2}) (P T_{Y_1})
$$

we get universal computing power by memo-izing the $P$ into

$$
S P T_{Y_1} T_{Y_2}
$$

What is interesting here conceptually is that we are obtaining totally
general pattern recognition capability, from the simple ingredients of

\begin{itemize}
\item
  Decision trees (i.e. single-feature queries and conditionals)
\item
  Recursion (mapping decision trees into bit strings and vice-versa)
\item
  Recognition of simple repeated patterns (i.e. the same $P$ is a pattern
  in both $T_{Y_2}$ and its argument $T_{Y_1}$)
\end{itemize}

This gives a novel perspective on the meaning and power of the $S$
combinator. $S$ is recognizing a simple repeated pattern. The universality
of $SK$ shows basically that all computable patterns can be built up from
simple repetitions -- if the "building up" involves recursion and
higher order functions.

In a phrase: {\it Distinction, If, Repetition-Recognition and Recursion Yield Universal Computation}.

This is nothing so new mathematically, but it's elegant conceptually to
thus interpret universality of $SK$ in terms of pattern recognition.

\section{Syntax-Semantics
Correlation}\label{syntax-semantics-correlation}

The line of thinking regarding decision trees, patterns and computations
presented above provides a new way of looking at syntax-semantics
correlation, which is a key concept in probabilistic evolutionary
learning \cite{Looks2006}and some other AI algorithms.

Syntax-semantics correlation means the correlation between two
distances:

\begin{itemize}
\item
  The distance between program P and program Q, in terms of the
  syntactic forms P and Q take in a particular programming language
\item
  The distance between P and Q, in terms of their manifestation as sets
  of input-output pairs
\end{itemize}

\noindent If this correlation is reasonably high, then syntactic manipulations can
be used as a proxy or guide to semantic manipulations, which can provide
significant efficiency gains.

It is known that, for the case of Boolean functions, it's possible to
achieve a relatively high level of syntax-semantics correlation in
relatively local regions of Boolean function space, if one adopts a
language that arranges Boolean functions in a certain hierarchical
format called Elegant Normal Form (ENF). ENF can be extended beyond
Boolean functions to general list operations and primitive recursive
functions, and this has been done in the OpenCog AI Engine in the
context of the MOSES probabilistic evolutionary learning algorithm.

The present considerations, however, suggest a more
information-theoretic approach to syntax-semantics correlation.

Consider first the case of Boolean functions. Suppose we re-organize a
Boolean function as a decision tree, choosing from among the smallest
such decision trees representing the function the one that does more
entropy-reduction toward the root of the tree (so one would rather have
the first decisions made while traversing the tree be the most
informative). Evaluating a pair of Boolean functions on a common
distribution of inputs, this should cause semantic and syntactic
distance to be fairly correlated.

Specifically, for the semantic distance between two functions f and g
defined on the same input space, consider the L1 distance evaluated
relative to a given probability distribution over inputs. For the
syntactic distance, consider first the decision-tree versions of f and
g, where each node is labeled with the amount of entropy reduction the
decision at that node provides. Then if one measures the edit distance
between the trees for f and g, but giving more cost to edits that
involve more highly-weighted (more entropy-reducing) nodes, one should
get a syntactic distance that correlates quite closely with semantic
distance. If one ignore the weights and gives more cost to edits that
occur higher up in the trees, one should obtain similar but weaker
correlation.

Of course, practical algorithms like XGBoost use greedy learning to form
decision trees and are thus crude approximations of the ``most
entropy-reducing among the smallest decision trees'' \ldots{} the trees
obtained from XGBoost don't actually give the optimal Huffman encoding,
so their relationship with entropy is only approximate. How good an
approximation the greedy approach will give in various circumstances is
difficult to say and requires context-specific analysis.

Going beyond Boolean functions, if one adopts the pattern-based SK model
described above, one has a situation where each pattern in a decision
tree T is equivalent to a decision tree on an input space consisting of
decision trees -- and one can think about the degree to which this
pattern increases or reduces entropy as it maps inputs into outputs.
Similar to the Boolean case, one can use an edit distance that weights
edits to more entropy-reducing nodes higher; or one can simplify and
weight more strongly those edits that are further from the tree leaves.

In this context, the rewriting done via Reduct rules in OpenCog today
becomes interpretable as a form of pattern recognition. The guideline
implied is that a Reduct rule should only be applied if there is some
reason to suspect it's serving as a pattern in the tree it's reducing.
I.e. for a rewrite rule \emph{source $\rightarrow$  target}, what we
want is that when running the rule backwards as \emph{target
$\leftarrow$ source}, the backwards rule constitutes a pattern in
\emph{source}. If this is the case, then the Reduct engine is carrying
out repeated acts of pattern recognition in a program tree, resulting in
a tree that has less informational redundancy than the initial version;
and likely there is higher syntax-semantics correlation across an
ensemble of such trees than among a corresponding ensemble of
non-reduced trees.

\section{Connecting Algorithmic and Statistical Complexity: Larger Higher-Order Decision Trees Have Higher Logical Entropy}\label{compactness-logical-entropy}

There are well known theorems relating algorithmic information to Shannon entropy; however these results have significant limitations.   It appears one can arrive at a less problematic relationship between information-theoretic uncertainty and compactness-of-expression by comparing {\it logical entropy} with {\it compactness of expression in the CoDD formalism}.

In the conventional algorithmic/Shannon case, if one has a source emanating bit strings according to a certain probability distribution, then for simple (low complexity) distributions the average Kolmogorov complexity of the generated bit strings is close to the Shannon entropy of the distribution; but these two quantities may be wide apart for distributions of high complexity.   

To be more precise, one can look at the average code-word length one would obtain by associating each bit-string with its maximally compressed representation as its code-word (where the average is calculated according to the assumed distribution over bit-strings); and then at the average code-word length one would obtain if one assigned more frequent bit-strings shorter code-words, which is roughly the entropy of the distribution.   The difference between these two average code-word lengths is bounded by the algorithmic information of the distribution itself (plus a constant).  \cite{GrunwaldVitanyi2010}.   

This is somewhat elegant, but on the other hand, complex probability distributions are the ones that we care most about in domains like biology, psychology and AI.   So it's also unsatisfying in a way.

A decision tree implies a partition of its input space, via associating each input with the partition defined by the path thru the decision tree that it follows.   It is then intuitive that, {\it on average} (roughly speaking -- I will make this more precise below) a random bigger decision tree will imply a partition w/ higher logical entropy than a random smaller decision tree.   The same intuitive reasoning applies to a random decision dag, or a random $k$'th order decision dag.    This is the CoDD incarnation of the intuition that ``bigger programs do higher entropy things."

The crux of the matter is the simple observation that {\it adding a decision node to a CoDD can increase the logical entropy of the partition the CoDD represents, but it can't decrease it}.

So if we measure the size of a CoDD as the number of decision nodes in it, then we know that {\it adding onto a CoDD will either increase the logical entropy or keep it constant}.   (Why would it be kept constant?  Basically if the added distinction made was then ignored by other distinctions intervening between it and the final output of the CoDD.)

If we view each step of adding a new decision node onto a CoDD as a random process, then on the whole larger CoDDs (which involve more steps to be added onto nothingness) are going to have higher logical entropy, as they involve more probably-logical-entropy-increasing expansion steps.  

So one concludes that, in the CoDD computational model

\begin{itemize}
\item adding onto a program does not decrease its logical entropy
\item on average bigger programs have higher logical entropy
\end{itemize}

\subsection{Connecting Algorithmic and Statistical Complexity in the Quantum Case}

What is the quantum version of this conclusion?   Baez \cite{baez2009physics} has presented an extension of classical combinatory logic that applies to the quantum case; so by considering these generalized combinators operating over quantum decision trees as described above, along with a  linear operator that flattens a quantum decision tree into a quantum state vector, one obtains a natural concept of a quantum CoDD.

The argument  becomes too involved to present here, but it seems to work out that adding a new decision node to a quantum CoDD cannot decrease the quantum logical entropy -- leading to the conclusion that a larger quantum CoDD will have greater quantum logical entropy.   Details of this case will be presented in a later paper.

\section{Conclusion}

Beginning from the foundational notion of distinction,  we have shown a new path to constructing and defining the concept of pattern, which has been used as the basis of theoretical analyses of general intelligence.   We have shown that the pattern concept thus formulated leads naturally to a novel  universal computational model, combinatory decision dags.  These CoDDs highlight subtle relationships between static program complexity and logical entropy, and runtime complexity and Shannon  entropy.  Further the key concepts appear to generalize to the quantum domain, potentially yielding elements of a future theory of quantum cognitive processes and structures.

Further work will apply  these concepts to the concrete analysis of particular classical and quantum cognitive processes, e.g.  in the context of evolutionary program learning systems, probability and amplitude based reasoning systems, and integrative cognitive architectures such as OpenCog.

\section*{Acknowledgments}

Conversations with Zar Goertzel were valuable in early stages of refining these ideas.   General  inspiration from Lou Kauffman and G. Spencer Brown's work on distinctions is  also  worth mentioning.

\bibliographystyle{splncs03} 
\bibliography{bbm}

\end{document}